# Deep learning-based virtual histology staining using auto-fluorescence of label-free tissue

Yair Rivenson[1,2,3,†], Hongda Wang[1,2,3,†], Zhensong Wei[1], Yibo Zhang[1,2,3], Harun Günaydın[1], Aydogan Ozcan[1,2,3,4,*]

[1]Electrical and Computer Engineering Department, University of California, Los Angeles, CA, 90095, USA

[2]Bioengineering Department, University of California, Los Angeles, CA, 90095, USA

[3]California NanoSystems Institute, University of California, Los Angeles, CA, 90095, USA

[4]Department of Surgery, David Geffen School of Medicine, University of California, Los Angeles, CA, 90095, USA.

†Equal contributing authors.

*Email: ozcan@ucla.edu

**ABSTRACT**

Histological analysis of tissue samples is one of the most widely used methods for disease diagnosis. After taking a sample from a patient, it goes through a lengthy and laborious preparation, which stains the tissue to visualize different histological features under a microscope. Here, we demonstrate a label-free approach to create a virtually-stained microscopic image using a single wide-field auto-fluorescence image of an unlabeled tissue sample, bypassing the standard histochemical staining process, saving time and cost. This method is based on deep learning, and uses a convolutional neural network trained using a generative adversarial network model to transform an auto-fluorescence image of an unlabeled tissue section into an image that is equivalent to the bright-field image of the stained-version of the same sample. We validated this method by successfully creating virtually-stained microscopic images of human tissue samples, including sections of salivary gland, thyroid, kidney, liver and lung tissue, also covering three different stains. This label-free virtual-staining method eliminates cumbersome and costly histochemical staining procedures, and would significantly simplify tissue preparation in pathology and histology fields.

Microscopic imaging of tissue samples is a fundamental tool used for the diagnosis of various diseases and forms the workhorse of pathology and biological sciences. The clinically-established gold standard image of a tissue section is the result of a laborious process, which includes the tissue specimen being formalin-fixed paraffin-embedded (FFPE), sectioned to thin slices (typically ~2-10μm), labeled/stained and mounted on a glass slide, which is then followed by its microscopic imaging using e.g., a bright-field microscope. All these steps use multiple reagents and introduce irreversible effects on the tissue. There have been recent efforts to change this workflow using different imaging modalities. One line of work imaged fresh, non-paraffin-embedded tissue samples using non-linear microscopy methods based on e.g., two-photon fluorescence, second harmonic generation[1], third-harmonic generation[2] as well as Raman scattering[3,4]. Another study used a controllable super-continuum source[5] to acquire multi-modal images for chemical analysis of fresh tissue samples. These methods require using ultra-fast lasers or super-continuum sources, which might not be readily available in most settings and require relatively long scanning times due to weaker optical signals. In addition to these, other microscopy methods for imaging non-sectioned tissue samples have also emerged by using UV-excitation on stained samples[6,7], or by taking advantage of the auto-fluorescence emission of biological tissue at short wavelengths[8]. In fact, auto-fluorescence signal creates some unique opportunities for imaging tissue samples by making use of the fluorescent light emitted from endogenous fluorophores. It has been demonstrated that such endogenous fluorescence signatures carry useful information that can be mapped to functional and structural properties of biological specimen and therefore have been used extensively for diagnostics and research purposes[8–10]. One of the main focus areas of these efforts has been the spectroscopic investigation of the relationship between different biological molecules and their structural properties under different conditions. Some of these well characterized biological constitutes include vitamins (e.g., vitamin A, riboflavin, thiamin), collagen, coenzymes, fatty acids, among others[9].

While some of the above discussed techniques have unique capabilities to discriminate e.g., cell types and sub-cellular components in tissue samples using various contrast mechanisms, pathologists as well as tumor classification software[11] are in general trained for examining histochemically stained tissue samples to make diagnostic decisions. Partially motivated by this, some of the above mentioned techniques were also augmented to create pseudo-Hematoxylin and Eosin (H&E) images[1,12], which were based on a linear approximation that relates the fluorescence intensity of an image to the dye concentration per tissue volume, using empirically determined constants that represent the mean spectral response of various dyes embedded in the tissue. These methods also used exogenous staining to enhance the fluorescence signal contrast in order to create virtual H&E images of tissue samples.

In this work, we demonstrate deep learning-based virtual histology staining using auto-fluorescence of unstained tissue, imaged with a wide-field fluorescence microscope through a standard near-UV excitation/emission filter set (see the Methods section). The virtual staining is performed on a single auto-fluorescence image of the sample by using a deep Convolutional Neural Network (CNN), which is trained using the concept of Generative Adversarial Networks (GAN)[13] to match the bright-field microscopic images of tissue samples after they are labeled with a certain histology stain (see Fig. 1 and Supplementary Figs. S1-S2). Therefore, using a

CNN, we replace the histochemical staining and bright-field imaging steps with the output of the trained neural net, which is fed with the auto-fluorescence image of the unstained tissue. The network inference is fast, taking e.g., ~0.59 sec using a standard desktop computer for an imaging field-of-view of ~ 0.33 mm × 0.33 mm using e.g., a 40× objective lens.

We demonstrated this deep learning-based virtual histology staining method by imaging label-free human tissue samples including salivary gland, thyroid, kidney, liver and lung, where the network output created equivalent images, very well matching the images of the same samples that were labeled with three different stains, i.e., H&E (salivary gland and thyroid), Jones stain (kidney) and Masson's Trichrome (liver and lung). Since the network's input image is captured by a conventional fluorescence microscope with a standard filter set, this approach has transformative potential to use unstained tissue samples for pathology and histology applications, entirely bypassing the histochemical staining process, saving time and cost. For example, for the histology stains that we learned to virtually stain in this work, each staining procedure of a tissue section on average takes ~45 min (H&E) and 2-3 hours (Masson's Trichrome and Jones stain), with an estimated cost, including labor, of $2-5[14,15] (H&E) and >$16-35[15,16] (Masson's Trichrome and Jones stain). Furthermore, some of these histochemical staining processes require time-sensitive steps, demanding the expert to monitor the process under a microscope, which makes the entire process not only lengthy and relatively costly, but also laborious. The presented method bypasses all these staining steps, and also allows the preservation of unlabeled tissue sections for later analysis, such as micro-marking of sub-regions of interest on the unstained tissue specimen that can be used for more advanced immunochemical and molecular analysis to facilitate e.g., customized therapies[17,18]. Also note that, this deep learning-based virtual histology staining framework can be broadly applied to other excitation wavelengths or fluorescence filter sets, as well as to other microscopy modalities (such as non-linear microscopy) that utilize additional endogenous contrast mechanisms. In our experiments, we used sectioned and fixed tissue samples to be able to provide meaningful comparisons to the results of the standard histochemical staining process. However, the presented approach would also work with non-fixed, non-sectioned tissue samples, potentially making it applicable to use in surgery rooms or at the site of a biopsy for rapid diagnosis or telepathology applications. Beyond its clinical applications, this method could broadly benefit histology field and its applications in life science research and education.

**RESULTS**

**Virtual staining of tissue samples**

We demonstrated the presented method using different combinations of tissue sections and stains. Following the training of a deep CNN (outlined in the Methods Section) we blindly tested its inference by feeding it with the auto-fluorescence images of label-free tissue sections that did not overlap with the images that were used in the training or validation sets. Figure 2 summarizes our results for a salivary gland tissue section, which was virtually stained to match H&E stained bright-field images of the same sample. These results demonstrate the capability of the presented framework to transform an auto-fluorescence image of a label-free tissue section into a bright-field equivalent image, showing the correct color scheme that is expected from an H&E stained tissue, containing various constituents such as epithelioid cells, cell nuclei, nucleoli, stroma, and collagen. An expert pathologist also confirmed that the neural network output images reported in Fig. 2 virtually stain and reveal the same histological features that are chemically stained. For example, a pathologist reported that both Figs. 2(c) and 2(d) confirm an irregular nest of atypical epithelioid cells in a background of desmoplastic stroma; cells also show variably-sized nuclei with mildly irregular contours and prominent nucleoli, and variable cells show smudgy chromatin in both the virtually stained image (Fig. 2(c)) and the chemically stained bright-field comparison (Fig. 2(d)). Similarly, Figs. 2(g) and 2(h) both confirm that a rounded nest of atypical epithelioid cells show retraction from the surrounding stroma and once again cells show variably-sized nuclei with mildly irregular contours and prominent nucleoli.

Next, we trained our deep network to virtually stain other tissue types with two different stains, i.e., the Jones stain (kidney) and the Masson's Trichrome stain (liver and lung). Figures 3 and 4 summarize our results for deep learning-based virtual staining of these tissue sections, which very well match to the bright-field images of the same samples, captured after the histochemical staining process. These results illustrate that the deep network is capable of inferring the staining patterns of different types of histology stains used for different tissue types, from a single auto-fluorescence image of a label-free specimen. With the same overall conclusion as in Fig. 2, it was also confirmed by a pathologist that the neural network output images Figs. 4(c, g) correctly reveal the histological features corresponding to hepatocytes, sinusoidal spaces, collagen and fat droplets (Fig. 4(g)), consistent with the way that they appear in the bright-field images of the same tissue samples, captured after the chemical staining (Figs. 4(d, h)). Similarly, it was also confirmed by the same expert that the deep network output images reported in Figs. 4(k, o) reveal consistently stained histological features corresponding to vessels, collagen and alveolar spaces as they appear in the bright-field images of the same tissue sample imaged after the chemical staining (Figs. 4(l, p)).

**Quantification of the network output image quality**

Next, beyond the visual comparison provided in Figs 2-4, we quantified the results of the deep network by first calculating the pixel-level differences between the bright-field images of the chemically stained samples and the virtually stained images that are synthesized using the deep neural network without the use of any labels/stains. Table 1 summarizes this comparison for different combinations of tissue types and stains, using the YCbCr color space, where the chroma components Cb and Cr entirely define the color, and Y defines the brightness component of the image. The results of this comparison reveal that the average difference between these two sets of images is < ~5% and < ~16%, for the chroma (Cb, Cr) and brightness (Y) channels, respectively. Next, we used a second metric to further quantify our comparison, i.e., the structural similarity index (SSIM)[19], which is in general used to predict the score that a human observer will give for an image, in comparison to a reference image. SSIM ranges between 0 and 1, where 1 defines the score for identical images. The results of this SSIM quantification are also summarized in Table 1, which very well illustrates the strong structural similarity between the network output images and the bright-field images of the chemically stained samples.

One should note that the bright-field images of the chemically stained tissue samples in fact do not provide the *true* gold standard for this specific comparison of the network output, because there are uncontrolled variations and structural changes (see e.g., Supplementary Fig. 3) that the tissue undergoes during the histochemical staining process and related dehydration and clearing steps. Another variation that we noticed for some of the images was that the automated microscope scanning software selected different auto-focusing planes for the two imaging modalities. All these variations create some challenges for the absolute quantitative comparison of the two sets of images (i.e., the network output for a label-free tissue vs. the bright-field image of the same tissue after the histological staining process). We further expand this point in the Discussion section.

**Transfer learning to other tissue-stain combinations**

Using the concept of transfer learning[20], the training procedure for new tissue and/or stain types can converge much faster, while also reaching an improved performance, i.e., a better local minimum in the training cost/loss function (see the Methods section). This means, a pre-learnt CNN model, from a different tissue-stain combination, can be used to initialize the deep network to statistically learn virtual staining of a new combination. Figure 5 demonstrates the favorable attributes of such an approach: a new deep neural network was trained to virtually stain the auto-fluorescence images of unstained *thyroid* tissue sections, and it was initialized using the weights and biases of another network that was previously trained for H&E virtual staining of the *salivary gland*. The evolution of the loss metric as a function of the number of iterations used in the training phase clearly demonstrates that the new thyroid deep network rapidly converges to a lower minimum in comparison to the same network architecture which was trained from scratch, using random initialization. Figure 5 also compares the output images of this thyroid network at different stages of its learning process, which further illustrates the impact of transfer learning to rapidly adapt the presented approach to new tissue/stain combinations. The network output images, after the training phase with e.g., ≥ 6,000 iterations, reveal that cell nuclei show irregular contours, nuclear grooves, and chromatin pallor, suggestive of papillary thyroid carcinoma; cells also show mild to moderate amounts of eosinophilic granular cytoplasm and the fibrovascular core at the network output image shows increased inflammatory cells including lymphocytes and plasma cells.

**DISCUSSION**

We demonstrated the ability to virtually stain label-free tissue sections, using a supervised deep learning technique that uses a single auto-fluorescence image of the sample as input, captured by a standard fluorescence microscope and filter set. This statistical learning-based method has the potential to restructure the clinical workflow in histopathology and can benefit from various imaging modalities such as fluorescence microscopy, non-linear microscopy, holographic microscopy and optical coherence tomography[21], among others, to potentially provide a digital alternative to the standard practice of histochemical staining of tissue samples. In this work, our method was demonstrated using fixed unstained tissue samples to provide a meaningful comparison to chemically stained tissue samples, which is essential to train the neural network as well as to blindly test the performance of the network output against the clinically approved method. However, the presented deep learning-based approach is broadly applicable to unsectioned, fresh tissue samples without the use of any labels or stains. Following its training, the deep network can be used to virtually stain the images of label-free fresh tissue samples, acquired using e.g., UV or deep UV excitation or even nonlinear microscopy modalities. Especially, Raman microscopy can provide very rich label-free biochemical signatures that can further enhance the effectiveness of the virtual staining that the neural network learns.

An important part of the training process involves matching the auto-fluorescence images of label-free tissue samples and their corresponding bright-field images after the histochemical staining process. One should note that during the staining process and related steps, some tissue constitutes can be lost or deformed in a way that will mislead the loss/cost function in the training phase (an example of this is illustrated in Supplementary Fig. S3). This, however, is only a training and validation related challenge and does *not* pose any limitations on the practice of a well-trained neural network for virtual staining of label-free tissue samples. To ensure the quality of the training and validation phase and minimize the impact of this challenge on the network's performance, we set a threshold for an acceptable correlation value between the two sets of images (i.e., before and after the histochemical staining process) and eliminated the non-matching image pairs from our training/validation set to make sure that the network learns the real signal, not the perturbations to the tissue morphology due to the chemical staining process. In fact, this process of cleaning the training/validation image data can be done iteratively: one can start with a rough elimination of the obviously altered samples (such as Supplementary Fig. 3), and accordingly converge on a neural network that is trained. After this initial training phase, the output images of each sample in the available image set can be screened against their corresponding bright-field images to set a more refined threshold to reject some additional images and further clean the training/validation image set. With a few iterations of this process, we can, not only further refine our image set, but also improve the performance of the final trained neural network.

We described above a methodology to mitigate some of the training challenges due to random loss of some tissue features after the histological staining process. In fact, this highlights another motivation to skip the laborious and costly procedures that are involved in histochemical staining as it will be easier to preserve the local tissue histology in a label-free method, without the need for an expert to handle some of the delicate procedures of the staining process, which sometimes also requires observing the tissue under a microscope.

The training phase of our deep neural network takes a considerable amount of time (e.g., ~13 hours for the salivary gland network) using a desktop PC; however, this entire process can be significantly accelerated by using dedicated hardware, based on GPUs. Furthermore, as already emphasized in Figure 5, transfer learning provides a warm start to the training phase of a new tissue/stain combination, making the entire process significantly faster. Unlike other color reconstruction or virtual staining approaches[12], once the deep network has been trained, the virtual staining of a new sample is performed in a single, non-iterative manner, which does not require a trial-and-error approach or any parameter tuning to achieve the optimal result. Based on its feed-forward and non-iterative architecture, the deep neural network rapidly outputs a virtually stained image in e.g., 0.59 sec, corresponding to a sample field-of-view of ~ 0.33 mm × 0.33 mm. With further GPU-based acceleration, it has the potential to achieve real-time performance, which might especially be useful in the operating room or for *in vivo* imaging applications.

The virtual staining procedure that is implemented in this work is based on training a separate CNN for each tissue/stain combination. If one feeds a CNN with the auto-fluorescence images of a different tissue/stain combination, it will not perform as desired (see e.g., Supplementary Fig. S4). This, however, is *not* a limitation because for histology applications, the tissue and stain type are pre-determined for each sample of interest, and therefore, a specific CNN selection for creating a virtually stained image from an auto-fluorescence image of the unlabeled sample does not require an additional information or resource. A more general CNN model can be learnt for multiple tissue/stain combinations by e.g., increasing the number of trained parameters in the model[22], at the cost of a possible increase in the training and inference times. Using a similar strategy, another avenue to explore in future work is the potential of the presented framework to perform multiple virtual stains on the same unlabeled tissue type.

As for the next steps, a wide-scale randomized clinical study would be needed to validate the diagnostic accuracy of the network output images, against the clinical gold standard, which will be important to better understand potential biases in the output images of the network. A significant advantage of the presented framework is that it is quite flexible: it can accommodate feedback to statistically mend its performance if a diagnostic failure is detected through a clinical comparison, by accordingly penalizing such failures as they are caught. This iterative training and transfer learning cycle, based on clinical evaluations of the performance of the network output, will help us optimize the robustness and clinical impact of the presented approach.

Finally, we would like to point to another exciting opportunity created by this framework for micro-guiding molecular analysis at the unstained tissue level, by locally identifying regions of interest based on virtual staining, and using this information to guide subsequent analysis of the tissue for e.g., micro-immunohistochemistry or sequencing[17]. This type of virtual micro-guidance on an unlabeled tissue sample can facilitate high-throughput identification of sub-types of diseases, also helping the development of customized therapies for patients[23].

**METHODS**

**Sample Preparation**

Formalin-fixed paraffin-embedded 2μm thick tissue sections were deparaffinized using Xylene and mounted on a standard glass slide using Cytoseal™ (Thermo-Fisher Scientific, Waltham, MA USA), followed by placing a coverslip (Fisherfinest, 24x50-1, Fisher Scientific, Pittsburgh, PA USA). Following the initial auto-fluorescence imaging process (using a DAPI excitation and emission filter set) of the unlabeled tissue sample, the slide was then put into Xylene for approximately 48 hours or until the coverslip can be removed without damaging the tissue. Once the coverslip is removed the slide was dipped (approximately 30 dips) in absolute alcohol, 95% alcohol and then washed in D.I. water for ~1 min. This step was followed by the corresponding staining procedures, used for H&E, Masson's Trichrome or Jones stains. This tissue processing path is only used for the training and validation of the approach and is *not needed* after the network has been trained. To test our method, we used different tissue and stain combinations: the salivary gland and thyroid tissue sections were stained with H&E, kidney tissue sections were stained with Jones stain, while the liver and lung tissue sections were stained with Masson's trichrome. The samples were obtained from the Translational Pathology Core Laboratory (TPCL) and were prepared by the Histology Lab at UCLA. All the samples were obtained after de-identification of the patient related information, and were prepared from existing specimen. Therefore, this work did not interfere with standard practices of care or sample collection procedures.

**Data acquisition**

The label-free tissue auto-fluorescence images were captured using a conventional fluorescence microscope (IX83, Olympus Corporation, Tokyo, Japan) equipped with a motorized stage, where the image acquisition process was controlled by MetaMorph® microscope automation software (Molecular Devices, LLC). The unstained tissue samples were excited with near UV light and imaged using a DAPI filter cube (OSFI3-DAPI-5060C, excitation wavelength 377nm/50nm bandwidth, emission wavelength 447nm/60nm bandwidth) with a 40×/0.95NA objective lens (Olympus UPLSAPO 40X2/0.95NA, WD0.65). Each auto-fluorescence image was captured with a scientific CMOS sensor (ORCA-flash4.0 v2, Hamamatsu Photonics K.K., Shizuoka Prefecture, Japan) with an exposure time of ~500 ms. The bright-field images (used for the training and validation) were acquired using a slide scanner microscope (Aperio AT, Leica Biosystems) using a 20×/0.75NA objective (Plan Apo), equipped with a 2× magnification adapter.

**Image pre-processing and alignment**

Since our deep neural network aims to learn a statistical transformation between an auto-fluorescence image of an unstained tissue and a bright-field image of the same tissue sample after the histochemical staining, it is of critical importance to accurately match the FOV of the input and target images. An overall scheme describing the global and local image registration process is described in Supplementary Fig. 5. The first step in this process is to find candidate features for matching unstained auto-fluorescence images and stained bright-field images. For this, each auto-fluorescence image (2048×2048 pixels) is down-sampled to match the effective pixel size of the bright-field microscope images. This results in a 1351×1351-pixel unstained auto-fluorescent tissue image, which is contrast enhanced by saturating the bottom 1% and the top 1% of all the pixel values, and contrast reversed to better represent the color map of the grayscale converted whole slide image (see Supplementary Fig. S5). Then, a normalized correlation score matrix is calculated by correlating each one of the 1351×1351-pixel patches with the corresponding patch of the same size, extracted from the whole slide gray-scale image. The entry in this matrix with the highest score represents the most likely matched FOV between the two imaging modalities. Using this information (which defines a pair of coordinates), we crop the matched FOV of the original whole slide bright-field image to create target images.

Following this FOV matching procedure, the auto-fluorescence and bright-field microscope images are coarsely matched. However, they are still not accurately registered at the individual pixel-level, due to the slight mismatch in the sample placement at the two different microscopic imaging experiments (auto-fluorescence, followed by bright-field), which randomly causes a slight rotation angle (e.g., ~1-2 degrees) between the input and target images of the same sample.

The second part of our input-target matching process involves a global registration step, which corrects for this slight rotation angle between the auto-fluorescence and bright-field images. This is done by extracting feature vectors (descriptors) and their corresponding locations from the image pairs, and matching the features by using the extracted descriptors[24]. Then, a transformation matrix corresponding to the matched pairs is found using the M-estimator Sample Consensus (MSAC) algorithm[25], which is a variant of the Random Sample Consensus (RANSAC) algorithm[26]. Finally, the angle-corrected image is obtained by applying this transformation matrix to the original bright-field microscope image patch. Following the application of this rotation, the images are further cropped by 100 pixels (50 pixels on each side) to accommodate for undefined pixel values at the image borders, due to the rotation angle correction.

Finally, for the local feature registration we applied an elastic image registration algorithm, which matches the local features of both sets of images (auto-fluorescence vs. bright-field), by hierarchically matching the corresponding blocks, from large to small (see Supplementary Fig. S5). The calculated transformation map from this step is finally applied to each bright-field image patch[27].

At the end of these registration steps, the auto-fluorescence image patches and their corresponding bright-field tissue image patches are accurately matched to each other and can be used as input and label pairs for the deep neural network training phase, allowing the network to *solely* focus on and learn the problem of virtual histological staining.

**Deep neural network architecture and training**

In this work, we used a GAN[13] architecture to learn the transformation from a label-free unstained auto-fluorescence input image to the corresponding bright-field image of the chemically stained sample. A standard convolutional neural network-based training learns to minimize a loss/cost function between the network's output and the target label. Thus, the choice of this loss function is a critical component of the deep network design. For instance, simply choosing an $\ell_2$-norm penalty as a cost function will tend to generate blurry results[28,29], as the network averages a weighted probability of all the plausible results; therefore, additional regularization terms[30,31] are generally needed to guide the network to preserve the desired sharp sample features at the network's output. GANs avoid this problem by learning a criterion that aims to accurately classify if the deep network's output image is real or fake (i.e., correct in its virtual staining or wrong). This makes the output images that are inconsistent with the desired labels not to be tolerated, which makes the loss function to be *adaptive* to the data and the desired task at hand. To achieve this goal, the GAN training procedure involves training of two different networks, as shown in Supplementary Figs. 1-2: (*i*) a *generator* network, which in our case aims to learn the statistical transformation between the unstained auto-fluorescence input images and the corresponding bright-field images of the same samples, after the histological staining process; and (*ii*) a *discriminator* network that learns how to discriminate between a true bright-field image of a stained tissue section and the generator network's output image. Ultimately, the desired result of this training process is a generator, which transforms an unstained auto-fluorescence input image into an image which will be *indistinguishable* from the stained bright-field image of the same sample. For this task, we defined the loss functions of the generator and discriminator as such:

$$\ell_{generator} = \text{MSE}\{z_{label}, z_{output}\} + \lambda \times \text{TV}\{z_{output}\} + \alpha \times (1 - D(z_{output}))^2$$
$$\ell_{discrimnator} = D(z_{output})^2 + (1 - D(z_{label}))^2 \qquad (1)$$

where $D$ refers to the discriminator network output, $z_{label}$ denotes the bright-field image of the chemically stained tissue, $z_{output}$ denotes the output of the generator network. The generator loss function balances the pixel-wise mean squared error (MSE) of the generator network output image with respect to its label, the total variation (TV) operator of the output image, and the discriminator network prediction of the output image, using the regularization parameters ($\lambda$, $\alpha$) that are empirically set to different values, which accommodate for ~2% and ~20% of the pixel-wise MSE loss and the combined generator loss ($\ell_{generator}$), respectively. The TV operator of an image $z$ is defined as:

$$\text{TV}(z) = \sum_{p}\sum_{q} \sqrt{(z_{p+1,q} - z_{p,q})^2 + (z_{p,q+1} - z_{p,q})^2} \qquad (2)$$

where $p$, $q$ are pixel indices. Based on Eq. (1), the discriminator attempts to minimize the output loss, while maximizing the probability of correctly classifying the real label (i.e., the bright-field image of the chemically stained tissue). Ideally, the discriminator network would aim to achieve $D(z_{label}) = 1$ and $D(z_{output}) = 0$, but if the generator is successfully trained by the GAN, $D(z_{output})$ will ideally converge to 0.5.

The generator deep neural network architecture follows the design of U-net[32], and is detailed in Supplementary Fig. S2. An input image is processed by the network in a multi-scale fashion, using down-sampling and up-sampling paths, helping the network to learn the virtual staining task at various different scales. The down-sampling path consists of four individual steps, with each step containing one residual block[33], each of which maps a feature map $x_k$ into feature map $x_{k+1}$:

$$x_{k+1} = x_k + \text{LReLU}\left[\text{CONV}_{k3}\left\{\text{LReLU}\left[\text{CONV}_{k2}\left\{\text{LReLU}\left[\text{CONV}_{k1}\{x_k\}\right]\right\}\right]\right\}\right] \qquad (3)$$

where CONV{.} is the convolution operator (which includes the bias terms), *k1*, *k2*, and *k3* denote the serial number of the convolution layers, and LReLU[.] is the non-linear activation function (i.e., a Leaky Rectified Linear Unit) that we used throughout the entire network, defined as:

$$\text{LReLU}(x) = \begin{cases} x & \text{for } x > 0 \\ 0.1x & \text{otherwise} \end{cases} \tag{4}$$

The number of the input channels for each level in the down-sampling path was set to: 1, 64, 128, 256, while the number of the output channels in the down-sampling path was set to: 64, 128, 256, 512. To avoid the dimension mismatch for each block[30], we zero-padded feature map $x_k$ to match the number of the channels in $x_{k+1}$ The connection between each down-sampling level is a 2×2 average pooling layer with a stride of 2 pixels that down-samples the feature maps by a factor of 4 (2-fold for in each direction). Following the output of the fourth down-sampling block, another convolutional layer maintains the number of the feature maps as 512, before connecting it to the up-sampling path.

The up-sampling path consists of four, symmetric, up-sampling steps, with each step containing one convolutional block. The convolutional block operation, which maps feature map $y_k$ into feature map $y_{k+1}$, is given by:

$$y_{k+1} = \text{LReLU}\left[\text{CONV}_{k6}\left\{\text{LReLU}\left[\text{CONV}_{k5}\left\{\text{LReLU}\left[\text{CONV}_{k4}\left\{\text{CONCAT}(x_{k+1}, \text{US}\{y_k\})\right\}\right]\right\}\right]\right\}\right] \tag{5}$$

where CONCAT(.) is the concatenation between two feature maps which merges the number of channels, US{.} is the up-sampling operator, and *k4, k5*, and *k6*, denote the serial number of the convolution layers. The number of the input channels for each level in the up-sampling path was set to 1024, 512, 256, 128 and the number of the output channels for each level in the up-sampling path was set to 256, 128, 64, 32, respectively. The last layer is a convolutional layer mapping 32 channels into 3 channels, represented by the YCbCr color map[34]. Both the generator and the discriminator networks were trained with a patch size of 256×256 pixels.

The discriminator network, summarized in Supplementary Fig. S2, receives 3 input channels, corresponding to the YCbCr color space of an input image. This input is then transformed into a 64-channel representation using a convolutional layer, which is followed by 5 blocks of the following operator:

$$z_{k+1} = \text{LReLU}\left[\text{CONV}_{k2}\left\{\text{LReLU}\left[\text{CONV}_{k1}\{z_k\}\right]\right\}\right] \tag{6}$$

where *k1, k2*, denote the serial number of the convolutional layer. The number of channels for each layer was 3, 64, 64, 128, 128, 256, 256, 512, 512, 1024, 1024, 2048. The next layer was an average pooling layer with a filter size that is equal to the patch size (256×256), which results in a vector with 2048 entries. The output of this average pooling layer is then fed into two fully connected layers with the following structure:

$$z_{k+1} = \text{FC}\left[\text{LReLU}\left[\text{FC}\{z_k\}\right]\right] \tag{7}$$

where FC represents the fully connected layer, with learnable weights and biases. The first fully connected layer outputs a vector with 2048 entries, while the second one outputs a scalar value. This scalar value is used as an input to a sigmoid activation function $D(z) = 1/(1+\exp(-z))$ which calculates the probability (between 0 and 1) of the discriminator network input to be real/genuine or fake, i.e., ideally $D(z_{label}) = 1$.

The convolution kernels throughout the GAN were set to be 3x3. These kernels were randomly initialized by using a truncated normal distribution[35] with a standard deviation of 0.05 and a mean of 0; all the network biases were initialized as 0. The learnable parameters are updated through the training stage of the deep network using an adaptive moment estimation (Adam) optimizer[36] with learning rate $1 \times 10^{-4}$ for the generator network and $1 \times 10^{-5}$ for the discriminator network. Also, for each iteration of the discriminator, there were 4 iterations of the generator network, to avoid training stagnation following a potential over-fit of the discriminator network to the labels. We have used a batch size of 10 in our training.

**Implementation details**

The other implementation details, including the number of trained patches, the number of epochs and the training times are shown in Table 2. The virtual staining network was implemented using Python version 3.5.0. The GAN was implemented using TensorFlow framework version 1.4.0. We implemented the software on a desktop computer with a Core i7-7700K CPU @ 4.2GHz (Intel) and 64GB of RAM, running a Windows 10 operating system (Microsoft). The network training and testing were performed using dual GeForce GTX 1080Ti GPUs (NVidia).

**Acknowledgements**

The Ozcan Research Group at UCLA acknowledges the support of NSF Engineering Research Center (ERC, PATHS-UP), the Army Research Office (ARO; W911NF-13-1-0419 and W911NF-13-1-0197), the ARO Life Sciences Division, the National Science Foundation (NSF) CBET Division Biophotonics Program, the NSF Emerging Frontiers in Research and Innovation (EFRI) Award, the NSF INSPIRE Award, NSF Partnerships for Innovation: Building Innovation Capacity (PFI:BIC) Program, the National Institutes of Health (NIH, R21EB023115), the Howard Hughes Medical Institute (HHMI), Vodafone Americas Foundation, the Mary Kay Foundation, and Steven & Alexandra Cohen Foundation. Yair Rivenson is partially supported by the European Union's Horizon 2020 research and innovation programme under the Marie Skłodowska-Curie grant agreement No H2020-MSCA-IF-2014-659595 (MCMQCT). The authors also acknowledge the Translational Pathology Core Laboratory (TPCL) and the Histology Lab at UCLA for their assistance with the sample preparation and staining, and Dr. Jonathan Zuckerman and Dr. Samuel French of UCLA Pathology Department for image evaluations.

**FIGURES AND TABLES**

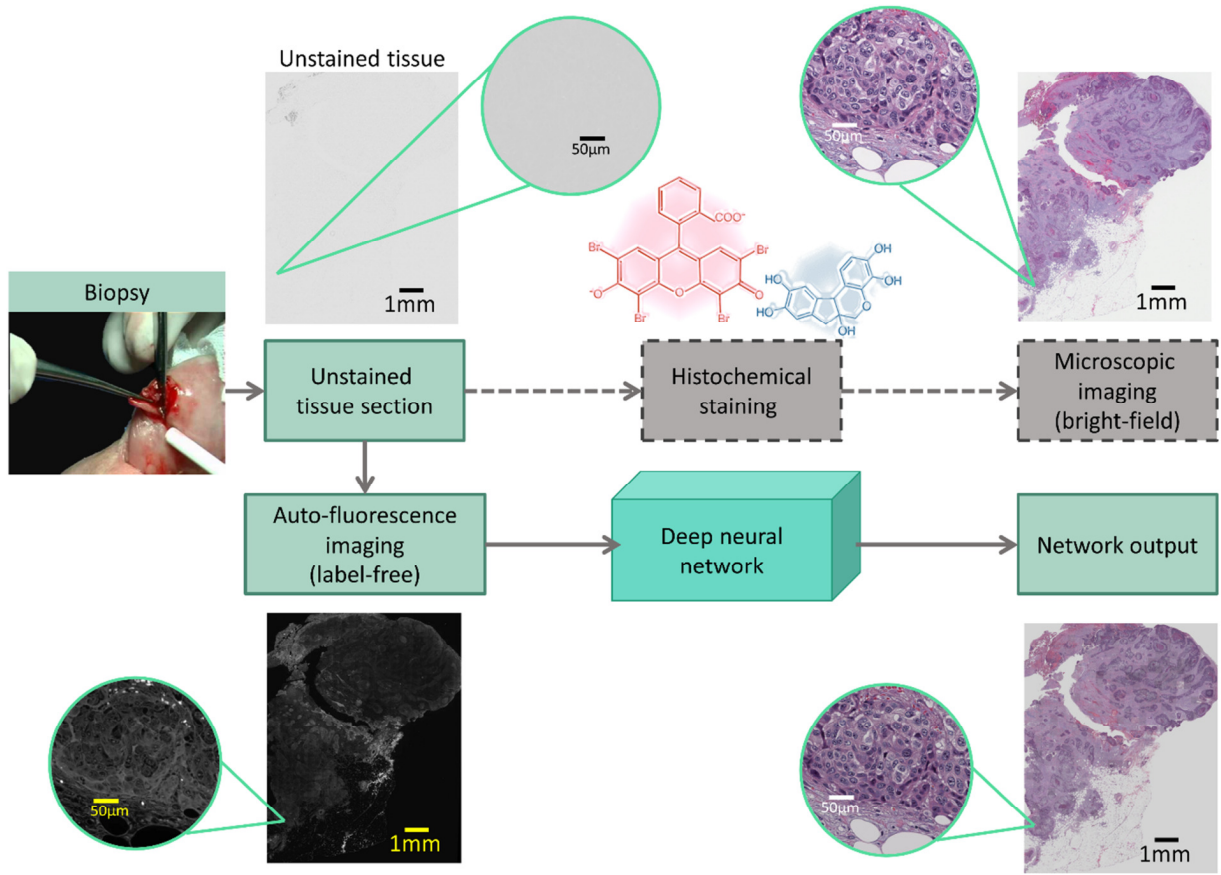

**Fig. 1**. Deep learning-based virtual histology staining using auto-fluorescence of unstained tissue. Following its training using a GAN, the neural network rapidly outputs a virtually stained tissue image, in response to an auto-fluorescence image of an unstained tissue section, bypassing the standard chemical staining procedure used in histology.

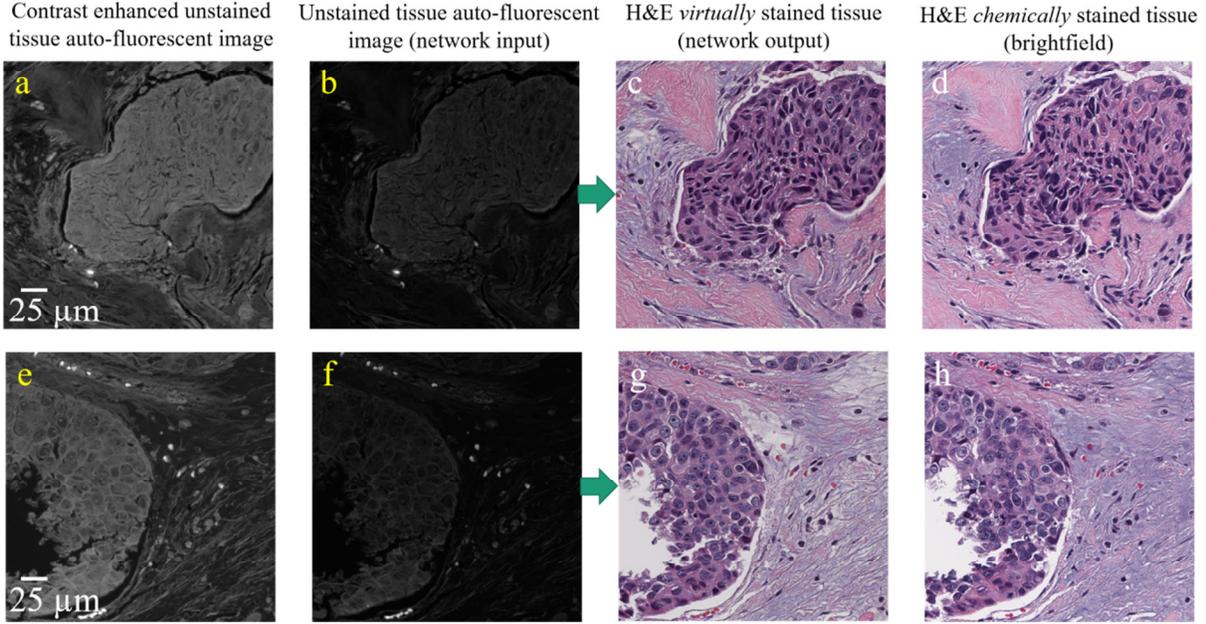

**Fig. 2**. Virtual staining results to match the H&E stain. The first 2 columns show the auto-fluorescence images of unstained salivary gland tissue sections (used as input to the neural network), and the third column shows our virtual staining results. The last column shows the bright-field images of the same tissue sections, after the histochemical staining process.

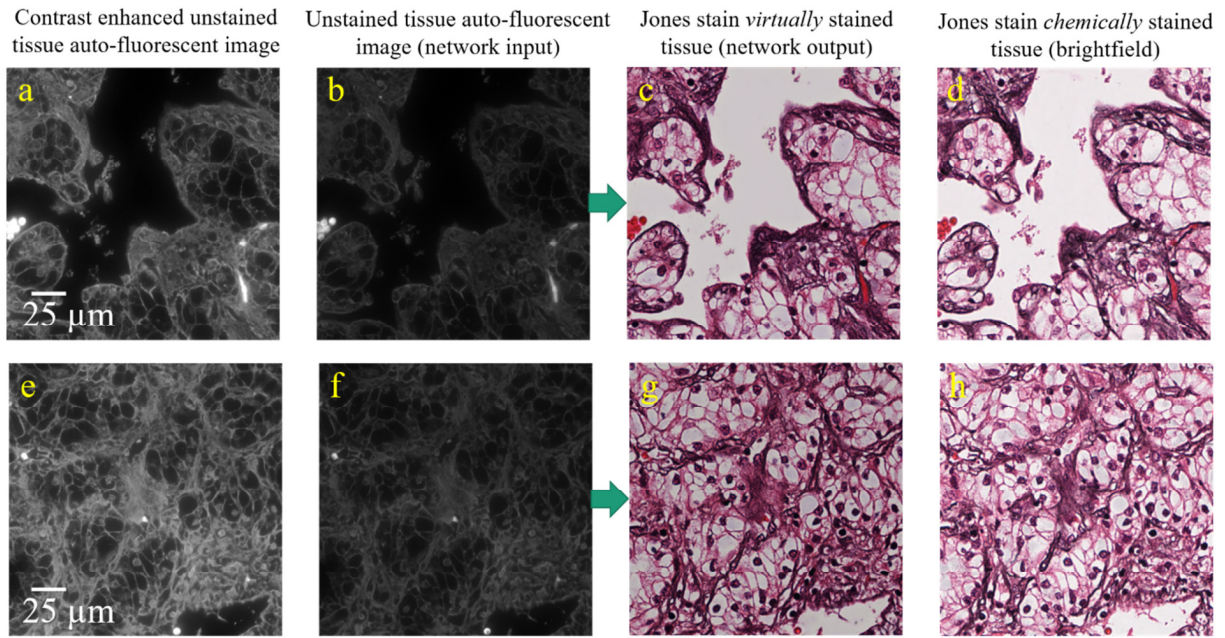

**Fig. 3**. Virtual staining results to match the Jones stain. The first 2 columns show the auto-fluorescence images of unstained kidney tissue sections (used as input to the neural network), and the third column shows our virtual staining results. The last column shows the bright-field images of the same tissue sections, after the histochemical staining process.

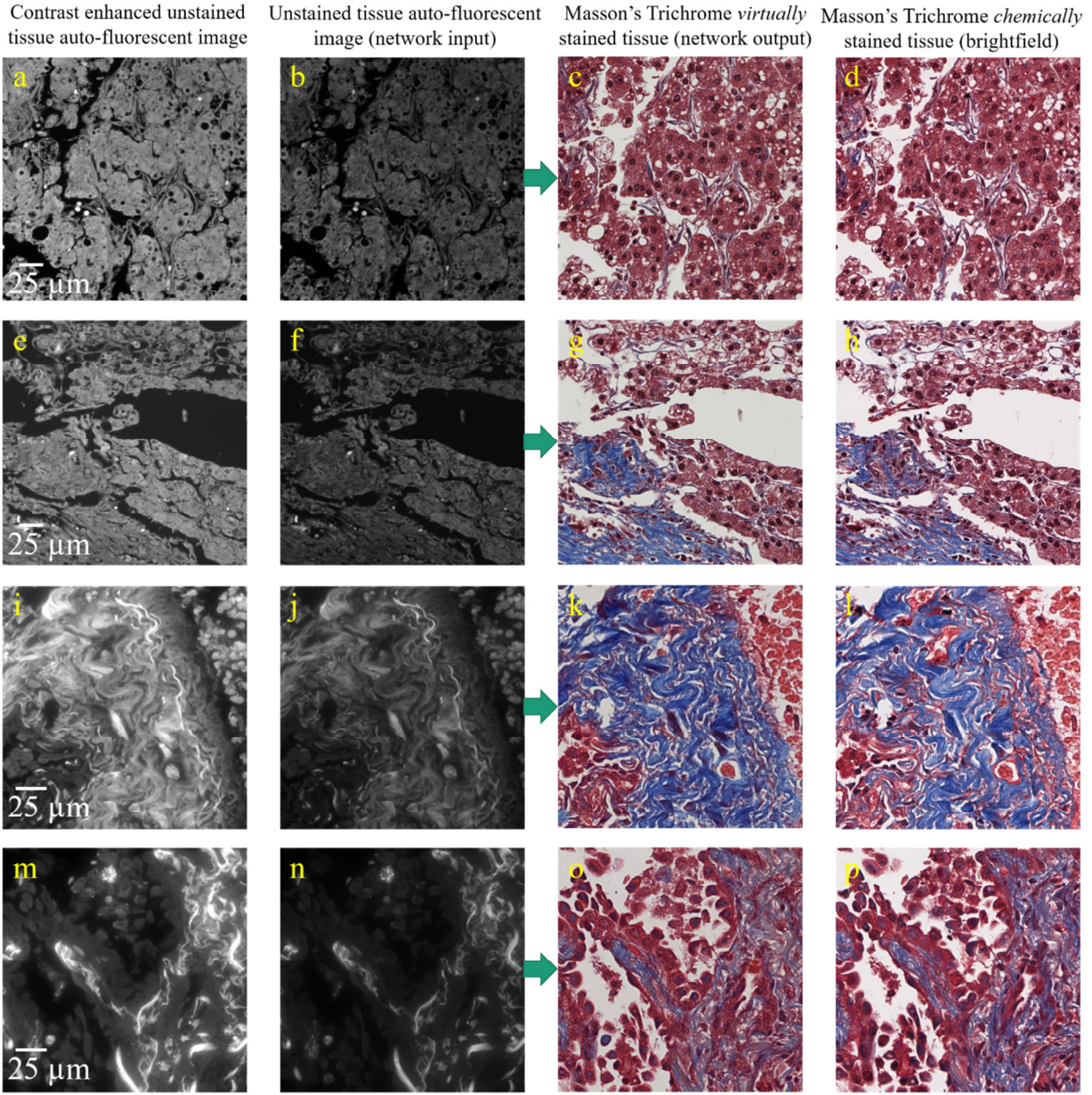

**Fig. 4**. Virtual staining results to match the Masson's Trichrome stain for liver and lung tissue sections. The first 2 columns show the auto-fluorescence images of an unstained liver tissue section (rows 1 and 2) and an unstained lung tissue section (rows 3 and 4), used as input to the neural network. The third column shows our virtual staining results for these tissue samples. The last column shows the bright-field images of the same tissue sections, after the histochemical staining process.

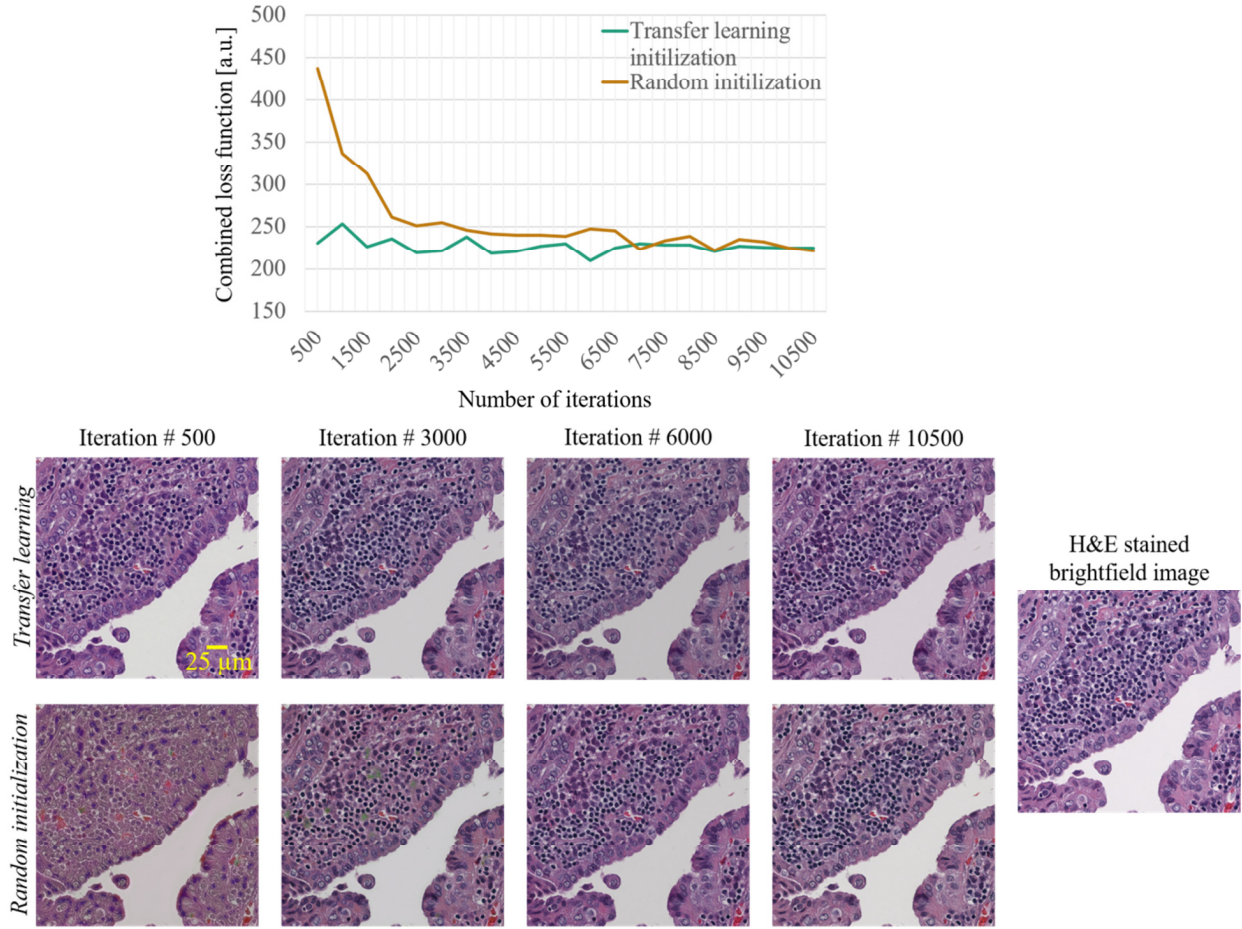

**Fig. 5**. Superior convergence is achieved using transfer learning. A new deep neural network is initialized using the weights and biases learned from the salivary gland tissue sections (see Fig. 2) to achieve virtual staining of thyroid tissue with H&E. Compared to random initialization, transfer learning enables much faster convergence, also achieving a lower local minimum. Network output images, at different stages of the learning process, are compared to each other to better illustrate the impact of the transfer learning to translate the presented approach to new tissue/stain combinations.

| Virtual histological staining using a deep network | Number of test images | SSIM | | Y difference (%) | | Cb difference (%) | | Cr difference (%) | |
|---|---|---|---|---|---|---|---|---|---|
| | | mean | std | mean | std | mean | std | mean | std |
| Salivary gland (H&E) | 10 | 0.826 | 0.059 | 11.5 | 9.0 | 2.5 | 2.4 | 2.5 | 2.5 |
| Thyroid (H&E) | 30 | 0.789 | 0.044 | 10.1 | 7.9 | 3.4 | 2.7 | 2.8 | 2.7 |
| Thyroid (H&E, transfer learning) | 30 | 0.839 | 0.029 | 14.0 | 8.4 | 2.4 | 2.2 | 2.6 | 2.6 |
| Liver (Masson's Trichrome) | 30 | 0.847 | 0.023 | 11.0 | 8.9 | 3.1 | 2.7 | 4.0 | 3.5 |
| Lung (Masson's Trichrome) | 48 | 0.776 | 0.039 | 15.9 | 11.7 | 4.0 | 3.6 | 5.3 | 4.9 |
| Kidney (Jones Stain) | 30 | 0.841 | 0.021 | 16.1 | 10.4 | 2.5 | 2.2 | 3.6 | 3.4 |

**Table 1**. Average and standard deviation (std) of the structural similarity index (SSIM) values as well as the brightness and chroma differences (defined by the YCbCr color space) are reported. These values are calculated between the network output images and the bright-field images of the same samples, captured after the histochemical staining. As detailed in the Discussion section, there is some unavoidable random loss or change of tissue features due to various steps used in the chemical staining process.

| Virtual staining network | # of training patches | # of epochs | Training time (hours) |
|---|---|---|---|
| Salivary gland (H&E) | 2768 | 26 | 13.046 |
| Thyroid (H&E) | 8336 | 8 | 12.445 |
| Thyroid (H&E, transfer learning) | 8336 | 4 | 7.107 |
| Liver (Masson's Trichrome) | 3840 | 26 | 18.384 |
| Lung (Masson's Trichrome) | 9162 | 10 | 16.602 |
| Kidney (Jones stain) | 4905 | 8 | 7.16 |

**Table 2**. Training details for different tissue/stain combinations.